\documentclass[10pt, conference, compsocconf]{IEEEtran}
\IEEEoverridecommandlockouts

\usepackage{amssymb, amsfonts}
\usepackage{booktabs}

\usepackage{cite}
\ifCLASSINFOpdf
   \usepackage[pdftex]{graphicx}
\else
\fi
\usepackage[cmex10]{amsmath}
\usepackage{algorithmic}
\usepackage[tight,footnotesize]{subfigure}
\begin{document}
\title{TSSRGCN: Temporal Spectral Spatial Retrieval \\Graph Convolutional Network for Traffic Flow Forecasting }


% \author{
% \IEEEauthorblockN{Authors Name/s per 1st Affiliation (Author)}
% \IEEEauthorblockA{line 1 (of Affiliation): dept. name of organization\\
% line 2: name of organization, acronyms acceptable\\
% line 3: City, Country\\
% line 4: Email: name@xyz.com}
% \and
% \IEEEauthorblockN{Authors Name/s per 2nd Affiliation (Author)}
% \IEEEauthorblockA{line 1 (of Affiliation): dept. name of organization\\
% line 2: name of organization, acronyms acceptable\\
% line 3: City, Country\\
% line 4: Email: name@xyz.com}
% \and 
% \IEEEauthorblockN{Authors Name/s per 1st Affiliation (Author)}
% \IEEEauthorblockA{line 1 (of Affiliation): dept. name of organization\\
% line 2: name of organization, acronyms acceptable\\
% line 3: City, Country\\
% line 4: Email: name@xyz.com}
% \and
% \IEEEauthorblockN{Authors Name/s per 2nd Affiliation (Author)}
% \IEEEauthorblockA{line 1 (of Affiliation): dept. name of organization\\
% line 2: name of organization, acronyms acceptable\\
% line 3: City, Country\\
% line 4: Email: name@xyz.com}

% }


\author{\IEEEauthorblockN{Xu Chen\IEEEauthorrefmark{1},
Yuanxing Zhang\IEEEauthorrefmark{1},
Lun Du\IEEEauthorrefmark{2}, 
Zheng Fang\IEEEauthorrefmark{1},
Yi Ren\IEEEauthorrefmark{1},
Kaigui Bian\IEEEauthorrefmark{1},
Kunqing Xie\IEEEauthorrefmark{1}\textsuperscript{$\S$} \thanks{$\S$ Corresponding author.}}
\IEEEauthorblockA{\IEEEauthorrefmark{1}School of Electrical and Computer Science, Peking University, Beijing, China,\\ 
Email: \{sylover, longo, fang\_z, renyi\_89, Bkg, kunqing\}@pku.edu.cn}
\IEEEauthorblockA{\IEEEauthorrefmark{2}Microsoft Research Asia, Beijing, China, \\Email: Lun.Du@microsoft.com}
}


\maketitle

\begin{abstract}

Traffic flow forecasting is of great significance for improving the efficiency of transportation systems and preventing emergencies. Due to the highly non-linearity and intricate evolutionary patterns of short-term and long-term traffic flow, existing methods often fail to take full advantage of spatial-temporal information, especially the various temporal patterns with different period shifting and the characteristics of road segments. Besides, the globality representing the absolute value of traffic status indicators and the locality representing the relative value have not been considered simultaneously. This paper proposes a neural network model that focuses on the globality and locality of traffic networks as well as the temporal patterns of traffic data. The cycle-based dilated deformable convolution block is designed to capture different time-varying trends on each node accurately. Our model can extract both global and local spatial information since we combine two graph convolutional network methods to learn the representations of nodes and edges. Experiments on two real-world datasets show that the model can scrutinize the spatial-temporal correlation of traffic data, and its performance is better than the compared state-of-the-art methods. Further analysis indicates that the locality and globality of the traffic networks are critical to traffic flow prediction and the proposed TSSRGCN model can adapt to the various temporal traffic patterns.
\end{abstract}


% \fi
\IEEEpeerreviewmaketitle

\section{Introduction}
With the expansion of human activities and the vigorous development of travel demands, transportation has become more and more important in our daily lives.
As a result, traffic flow forecasting has attracted the attention of government agencies, researchers, and individual travelers. 
%The accurate predictions enable the transport department to regulate traffic and prevent congestion, as well as help travelers plan to travel through uncongested roads. 
Predicting future traffic flows is nowadays one of the critical issues for intelligent transportation systems (ITS), and becomes a cutting-edge research problem.
% The past few years have witnessed the development of ITS.
With the deployment of more traffic sensors, a large amount of real-time traffic data can be easily collected for scientific study.
A challenging issue for a practical ITS is to recognize the evolutional patterns through the massive data from the sensors.
% However, it is challenging to find evolution patterns in such rich data over the long term.
Legacy solutions \cite{arima,svr} provide essential solutions, while they cannot capture spatial and temporal dependency concurrently.
%leading to the degradation of accuracy on the forecasting task.
Models based on recurrent neural networks (RNNs) have made significant progress on this issue, 
yet it may be challenging to learn mixture periodic patterns within the collected data.
A recent study directly divides raw data into weekly/daily/recent data sources \cite{astgcn} as the manual supervision to mine temporal features. 
% under the sacrifice of the flexibility of the evolution patterns.
% , yet it may lose adequate information about evolution patterns as it treats them in the same way. 
However, the temporal cycle of traffic may not be constant due to occasions or other factors like climate and interim regulations.
% fluctuate slightly due to climate and other factors, increasing the difficulty of traffic prediction.
This arises the first critical research question (RQ) for traffic flow prediction: \emph{RQ1: How to design a module to dynamically capture various temporal patterns of traffic data?}

% Apart from the temporal aspects of traffic, 
The traffic flow forecasting task also faces challenges from the spatial aspects.
% First, the most intuitive but crucial challenge is: \emph{How to further improve the traffic flow forecasting in the dynamic traffic network?}
The previous effort mainly focuses on globality (i.e., the absolute value of traffic flow) of sensors while ignoring investigation of locality (i.e., the relative value compared to upstream or downstream sensors).
Usually, locality of sensors provides evidence for a snapshot of traffic flow in the near future.
%Fig.~\ref{fig:intro}(a) illustrates a common phenomenon in real-world traffic. 
Considering two road segments $A_1\rightarrow B_1$ and $A_2\rightarrow B_2$ under the same traffic condition at timestep $t_0$. Globality of sensor $B_1$ and $B_2$ are the same in the beginning, while more cars passing $A_1$ and less passing $A_2$, resulting in a significant increase in the flow near $B_1$ and a remarkable decrease near $B_2$.
% The following example further illustrates this phenomenon.
% As shown in Fig.~\ref{fig:intro}(a), there are two pairs of sensors $A_1\to B_1$ and $A_2\to B_2$ that represent two road segments both with one-way lanes. 
% The globality of the sensor $B_1$ and $B_2$, which means the absolute value of traffic flow, are equal at timestep $t_0$.
% As time goes by, the locality representing the relative value between the upstream sensor and itself will have a significant influence on its globality, as cars passing sensors $A_1$ and $A_2$ will move to their downstream sensors, 
This demonstrates the importance of the correlation between neighboring sensors, yielding the second research question: \emph{RQ2: How to learn and use graph structures to adequately describe local and global features of a transportation network?}

Rethinking the locality of sensors, we discover that it reflects the relative relation of the traffic status between neighboring nodes. 
Intuitively, it is natural to take road segments into consideration for characterizing the locality.
Owing to the particular geographical location and characteristics of each road, traffic flow tends to present various patterns on different roads.
%Fig.~\ref{fig:intro}(b) illustrates an example of 
Assume that there are two road segments $A_3\to B_3$ and $A_4\to B_4$ with different number of lanes, i.e. the former is a one-lane road while the latter has double amount of lanes.
% Should there be the same number of cars on these two road segments? 
The number of cars on these two road segments is the same at time step $t_0$. With higher capacity, $A_4\to B_4$ can accommodate more cars at a high speed at $t_0+\Delta t$.
% Apparently, cars on $A_4\to B_4$ tend to drive faster as the capacity of $A_4\to B_4$ is higher than $A_3\to B_3$, causing more cars in the former road segment at future time step $t_0+\Delta t$.
However, it is difficult and expensive to obtain the explicit and exact description of intrinsic characteristics and instantaneous states towards all roads.
Therefore, the third research question comes down to: \emph{RQ3: How to incorporate the above information through embedding edges for better predicting traffic flow of nodes?}

The advancement of Graph Convolutional Networks (GCNs) \cite{gcn} introduces many variants to capture spatial correlations, boosting the prosperity of modeling traffic networks as graphs. 
Enlightened by the promising performance of GCNs on many graph-based inference tasks, in this paper, we propose a novel traffic flow forecasting model, named \underline{T}emporal \underline{S}pectral \underline{S}patial \underline{R}etrieval \underline{G}raph \underline{C}onvolutional \underline{N}etwork (TSSRGCN), to address the above RQs. 
In TSSRGCN, a cycle-based dilated deformable convolution block is employed to introduce prior background knowledge into the model to mine meaningful temporal patterns and expand the receptive field in the time dimension (for RQ1).
We then involve a Spectral Spatial Retrieval Graph Convolutional block comprising a Spectral Retrieval layer and a Spatial Retrieval layer to model the locality and globality of the traffic network from the perspective of spatial dimension (for RQ2).
Meanwhile, the edges are transformed into representations by the exploitation of the connected nodes over a specific period (for RQ3).
Our model is capable of capturing spatial-temporal correlations and is sufficient for time-varying graph-structured data.
Evaluations over two real-world traffic datasets verify that the proposed TSSRGCN outperforms the state-of-the-art algorithms on different metrics.

The main contribution of this paper is summarized as follow:
1. We reconsider the character of different temporal patterns and adopt dilated convolution as well as deformable convolution for mining useful traffic evolution patterns. 
 The reasonable period is concerned to precisely capture the time-varying pattern. Besides, the period shifting of each pattern is also considered and learned in the well-designed block. 
2.The spectral spatial retrieval graph convolutional block is proposed to extract the geographical structure of the traffic network from global and local perspectives. 
    Unlike traditional graph convolution methods, the edge information is considered in this block to build the spatial correlation between nodes and edges. 
 3.We achieve state-of-the-art performance by evaluating our model on two real-world datasets. In-depth analyses show that the design of TSSRGCN enhances the robustness and effectiveness under various traffic patterns.
%\begin{itemize}
%    \item 
%    \item     
%    \item 
%\end{itemize}

\begin{figure*}[tp]
	\centering
	\includegraphics[width=1.0\textwidth]{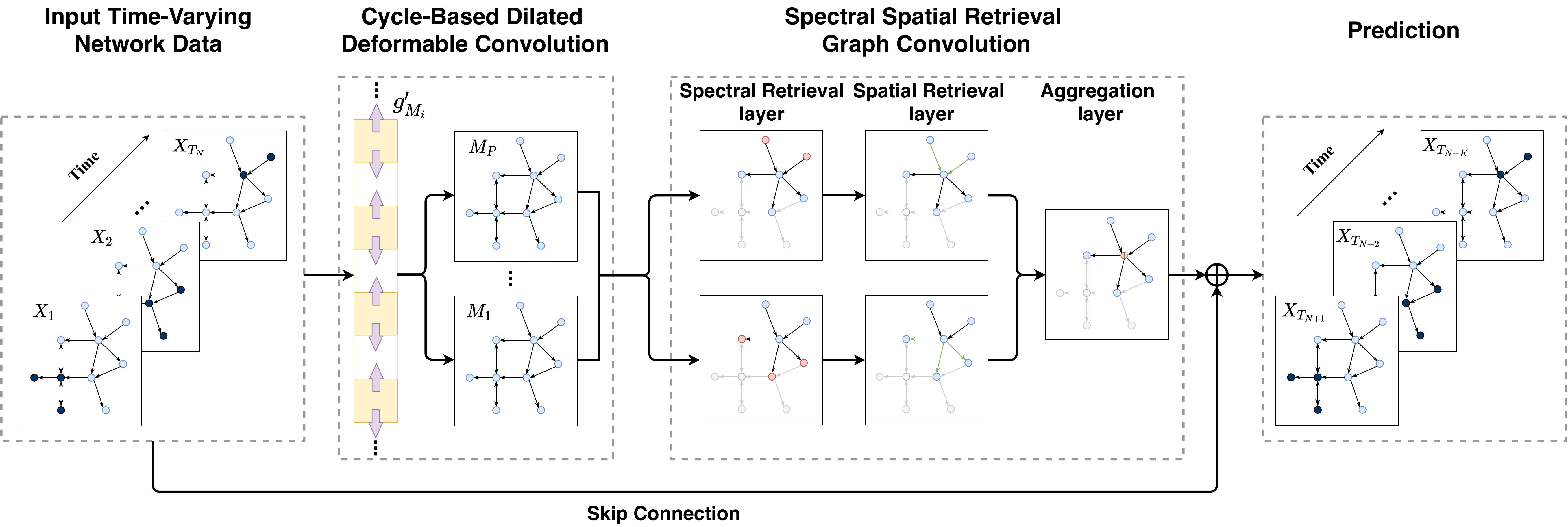}
	\caption{The framework of TSSRGCN. Cycle-Based Dilated Deformable Convolution block with kernel $g'_{M_i}$ is firstly applied to capture $P$ temporal patterns with dilated rate $M_i,i\in\{1,\ldots,P\}$. Stacked Spectral Spatial Retrieval Graph Convolution blocks model nodes and edges by extracting the locality and globality of the entire traffic network. Skip connection is utilized to concatenate input and embedding before the last fully-connected layer. The model is able to generate accurate traffic flow prediction from short term to long term.}
	\label{fig:framework}
\end{figure*}

\section{Preliminaries}
% \subsection{Traffic Network}
Denote the traffic network as a directed weighted graph $G=(V, E, W^0, W^1)$ in this paper. 
Here, $V$ is the set of nodes with size $N=|V|$ and $E$ is the set of edges representing the road segments between sensors. 
% $W \in \mathbb{R}^{N\times N\times C} $ is the weighted adjacency matrix constructed based on network structure where $C$ denotes the number of types of weighted adjacency matrix such as $0-1$ adjacency matrix or distance matrix.
$W^0 \in \{0,1\}^{N\times N}$ and $W^1 \in \mathbb{R}^{N\times N} $ are the general and weighted adjacency matrices respectively constructed based on the graph.

% Sensors are placed on each node to measure and record variable indicators of traffic network with setting sampling frequency $T_{sp}$. 
Sensors periodically measure and record traffic status indicators such as flows, occupancy, and speed.
Let $T_{sp}$ denote the measuring frequency and $F$ denote the number of the recorded indicators.
Given a time interval $[T_{begin}, T_{end}]$, the traces of indicators recorded by $N$ sensors can be represented by $X\in\mathbb{R}^{N\times F\times T_N}$.
$T_N=(T_{end}-T_{begin})/T_{sp}$ stands for the number of timesteps in the given interval.
%The data generated by $N$ sensors during time interval $[T_{begin},T_{end}]$ are $X$ with shape $\mathbb{R}^{N\times T_N\times F}$. 
% $T_N=(T_{end}-T_{begin})/T_{sp}$ means number of timesteps in $[T_{begin},T_{end}]$ and $F$ denotes the number of indicators pre-designed in the traffic sensor system. 

% \subsection{Traffic Flow Forecasting Problem}
The traffic flow forecasting task on the network aims to provide accurate predictions of the future flow, which can be formulated as:
Given an traffic network $G=(V, E, W^0, W^1)$ and historical data $X=(X_{1},\ldots,X_{T_N})$ where $X_{T_i} \in \mathbb{R}^{N\times F}$ is data at timestep $T_i$, we are expected to learn a function $\Psi$ for prediction of traffic flow series $\hat{Y}=\Psi(X)=(\hat{Y}_{T_{N+1}},\ldots,\hat{Y}_{T_{N+K}})$ for all nodes in the next $K$ time steps after $T_N$, i.e.,:
\begin{equation}
%     \hat{X}_{T_{N+1}},\ldots,\hat{X}_{T_{N+K}}= \Psi(X_{T_1},\ldots,X_{T_N};G) 
    \Psi^* = \arg\min_{\Psi} ||\Psi(X) - Y||^2,
\end{equation}
Here $Y_{T_i}\in \mathbb{R}^{N}$ is the ground-truth flow at $T_i$ and $Y=(Y_{T_{N+1}},\ldots,Y_{T_{N+K}})$ is the time series to be forecasted.

% \begin{equation}
% \begin{split}
% & \hat{X}_{T_{N+1}},\ldots,\hat{X}_{T_{N+K}}=  \\
% & \mathop{\arg\max}_{X_{T_{N+1}}, \ldots,X_{T_{N+K}}} \log \mathrm{P}(X_{T_{N+1}}, \ldots, X_{T_{N+K}}|X;G) 
% \end{split}
% \end{equation}

\section{Temporal Spectral Spatial Retrieval Graph Convolutional Network}

In this paper, we propose Temporal Spectral Spatial Retrieval Graph Convolution Network (TSSRGCN) for accurate traffic flow forecasting.
The overall architecture of the proposed TSSRGCN is illustrated in Fig.~\ref{fig:framework}. 
TSSRGCN consists of a cycle-based dilated deformable convolution block (CBDDC-block), stacked spectral spatial retrieval graph convolutional block (SSRGC-block) and a fully-connected layer for the final prediction. 
%CBDDC-block is composed of $P$ different convolution operators to capture different temporal evolution patterns (RQ1).
%SSRGC-block contains a Spectral Retrieval layer, a Spatial Retrieval layer, an Aggregation layer and a 2-D CNN layer for spatial information mining (RQ2 and RQ3). 
Skip connection \cite{resnet} is applied to fuse the high-order knowledge learned from the stacked SSRGC-blocks and the low-order input features. 
% Similarly, the residual connection is employed to concatenate the embedding of sensors and the input to increase the depth of SSRGC-block.
%We add a fully-connected layer in the last layer to yield the traffic flow prediction for each sensor.
The detailed design of TSSRGCN will be explained in this section.

\subsection{Cycle-Based Dilated Deformable Convolution Block}
\noindent\textbf{Cycle-Based Dilated Convolution}.
%\noindent\textbf{Standard Dilated Convolution}.
%We revisit dilated convolution and adapt it for temporal pattern recognition in TSSRGCN. 
Dilated convolution is proposed by \cite{dilated_nn} in computer vision to exponentially expand the receptive field without loss of resolution.
\cite{graphwavenet} adopts this in convolution block to learn temporal features of nodes.
%time-varying trends of nodes. 
%Dilated convolution is more efficient than RNN-based models in handling time series as the parallel computation of convolution operation. 
Different from the vanilla convolution, kernels of dilated convolution are sparse as the dilated rate is usually set to a power of two, i.e. $dr=2^i,i\in \mathbb{N}^+$, with $dr$-1 pixels skipped. 
%A low dilation rate leads to a concentration on short-distance information, while a large dilation rate brings investigation on long-distance information.

%As illustrated by Figure, there are $dr$-1 zeros between two learnable elements of the kernels. \ref{fig:without}. 
Considering a series of traffic data collected in $T_N$ days and the dilated rate is $dr=2^2$, then it will extract information from data at day $1,5,\ldots,T_{4\times \lceil T_N/4 \rceil-3}$, which maybe Monday, Friday, Next Tuesday, etc. 
The selected days does not form a regular traffic period, meaning the convolution receives a meaningless input series and thereby fails to retrieve knowledge of traffic patterns.

Intuitively, the traffic data may present three temporal evolution patterns \cite{astgcn}: daily/recent/weekly. 
The \emph{daily} pattern implies that the temporal trend in every two adjacent days may be very similar, while \emph{Recent} traffic status has a strong impact on current timestep for it is likely to continue the trend of previous timestep.
%交代三种模式的特点 
The \emph{weekly} pattern has two effects: the same moment in different weeks show a similar basic pattern, named discrete weekly periodicity; the traffic status continuously evolves from last week to this week due to various factors, named the continuous weekly trend. Existing research only focuses on weekly periodicity while ignores weekly trends, losing rich evolution information. 

To incorporate the above temporal patterns, we propose a cycle-based dilated convolution block. The dilated rate is restricted to be chosen from a pre-define dilated rate set $\tau$.
The daily pattern and two weekly evolution effects can be described in $\tau$ depending on the data sampling frequency $T_{sp}$. Note that more periods like monthly and seasonal one can be further defined with more data and computing resources.
% Besides, dilated convolution always begins with a lower dilated rate and the receptive filed is expanded by stacking layers, which may lead to vanishing gradient or gradient explosion and it is difficult for the model to converge. 
Given a period length (i.e., dilated rate) $M_i\in\tau$, the corresponding cycle-based dilated convolution on input $x_{s}(t)\in X_t$ of node $s$ at time step $t$ is depicted as
    \begin{equation}\label{eqn:CBDC}
        (g_{M_i}  \star_\tau x_s  )(t) = \sum^{K_S(M_i)-1}_{p=0} g_{M_i}(p)x_s(t- M_i \times p)
    \end{equation}
where $g_{M_i} \in \mathbb{R}^{K_S(M_i)}$  
is a dilated convolution kernel containing $K_S(M_i)$ elements. The element of $g_{M_i}$ is indexed by $p$. 
% and $x^s(t-M_i\times l)$ is feature of input data at time step $t-M_i\times l$. 
% $p$ denotes the $p$-th element of the kernel, 
$\star_\tau$ denotes the cycle-based dilated convolution over $\tau$.

% Through CBDDC-block, weekly periodicity can be directly captured while weekly trend will be extracted by learning daily overview of traffic conditions and then embedding it into the weekly trend representation. 
% Daily pattern will focus on each time step during the given time of a day.

\noindent\textbf{Cycle-Based Dilated Deformable Convolution}.
 The temporal dynamics could have some perturbation between two periods \cite{revisiting};
% , so there may exist the temporal periodicity shifting. 
thus, fixed periods in $\tau$ ignore the period shifting, leading to biased learning.
% Enlightened by \cite{deformable}, w
We adapt the deformable convolution \cite{deformable} to tackle this problem. 
% Deformable convolution is proposed to solve the problem in the area of computer vision that CNN with fixed geometric structure is weak at capturing different geometric transformations. 
% In the traffic scene, b
By adding learnable position shiftings to the kernels, convolution operation could adaptively represent various temporal patterns and be flexible to capture the variation of periods.
%The position shiftings are set as learnable parameters, 
%% As the position shiftings are learnable, 
%with which the block can adaptively represent the various temporal pattern.
%% with their period shifting.
The Cycle-Based Dilated Deformable Convolution block modifies Eqn.~(\ref{eqn:CBDC}) by:
    \begin{equation}
        (g'_{M_i}  \star_\tau x_s  )(t) = \sum^{K_S(M_i)-1}_{p=0} g'_{M_i}(p)x_s(t- M_i \times p + \Delta p)
    \end{equation}
with $g'_{M_i} \in \mathbb{R}^{K_S(M_i)}$ as dilated deformable convolution kernel. $\Delta p$ is the position shifting for $p$-th kernel element.
% that is the key to capture the period shifting. 
% and $x^s(t-M_i\times l)$ is feature of input data at time step $t-M_i\times l$.  

%In practice, CBDDC-block portrays the daily pattern over each time step throughout a given duration of several days. The weekly periodicity is captured similarly.
%Regarding the weekly trend, CBDDC-block forms the daily overview of traffic conditions on each node by a simple CNN, then extracts the trend across daily overview and embeds the knowledge into weekly trend representation.
In practice, we apply $P$ dilated deformable convolution layers
%which also contains a layer with $dr=1$ conditioned on the nearest 12 time steps to extract the latest information. 
to ensures that TSSRGCN would depict the temporal aspects of traffic networks over dynamic evolution patterns.
The outputs of all layers $\mathcal{P}_i,i\in\{1,\ldots,P\}$ will be concatenated and fed to a linear layer with learnable parameter $\Omega\in\mathbb{R}^{P\times F\times F_T}$ for feature fusion as the \emph{temporal representation} for each node.
The temporal representation is then taken as the input to the next block, denoted as $H_0= [\mathcal{P}_1,...,\mathcal{P}_P]\Omega$,
where $[,]$ is a concatenate operation and $F_T$ denotes the dimention of the temporal representation.
In this case, the temporal aspect of traffic patterns is retrieved by the CBDDC layers with various dilated rates (i.e. CBDDC-block).

% Compared with traditional dilated convolution, we directly choose meaningful dilated rate $p_i$ like $p_i=8$ to capture weekly pattern on time dimension as the PBDDC layer extracts data at the same timestamp on different days. Moreover, it also helps to increase the receptive field of the convolution operation because of the larger dilated rate on the first layer. Following this idea, daily pattern could be also obtained by setting $p_i=1$. 

\subsection{Spectral Spatial Retrieval Graph Convolutional Block}

TSSRGCN employs the Spatial Spectral Retrieval Graph Convolutional block (SSRGC-block) to investigate the sensors data and features of road segments on traffic network.
% The locality and globality concerns are expected to be integrated into the model. 
A SSRGC-block is composed of a spectral retrieval layer, a spatial retrieval layer and an aggregation layer. 
Specifically, the spectral retrieval layer is applied to aggregate information from upstream and downstream nodes based on spectral methods, respectively. 
The node embedding learned by the spectral retrieval layer will be used to generate edge representations through the spatial retrieval layer with edge information from the weighted adjacency matrix. 
%The SSRGC-block integrates spectral-based GCN and spatial-based GCN to capture the information from both nodes and edges, and thereby it can depict the locality and globality of the traffic network.
An aggregation operation finally aggregates edge representations to the connected nodes to retrieve the critical evidence for the forecasting. 
In practice,  we stack $\lambda$ blocks in TSSRGCN for consideration of efficiency and accuracy. 

\noindent\textbf{Spectral Retrieval Layer}. 
% GCN is proposed to learn node embedding from its neighborhood and its feature. 
Spectral-based GCN utilizes adjacency matrix $W^0$
% or Laplace matrix $L=D_d-W$ 
to aggregate information from neighborhood nodes. % where $D_d=\text{diag}(W\textbf{1})$ is out-degree matrix of the traffic network. 
For example, 
%\cite{kipf2016semi} proposes a first-order approximation of spectral graph convolution by the truncated expansion of Chebyshev polynomials. 
%To model the spatial-temporal correlation,
 \cite{dcrnn} proposes a diffusion convolution layer with the motivation that spatial structure in traffic is non-Euclidean and directional; thus, upstream and downstream nodes can have different influence on current nodes. 
We adapt the diffusion convolution layer to spread the messages over two directions. 
% Given the traffic graph, 
Let $A_d=D_d^{-1}W^0$ denote the transition matrix, which measures the probability that information of current node transfers to its downstream neighbor,
where $D_d=\text{diag}(W^0\textbf{1})$ is out-degree matrix.
Similarly, $A_u=D_u^{-1}{W^0}^\mathsf{T}$ could be used to gather information from upstream neighborhood with $D_u=\text{diag}({W^0}^\mathsf{T}\textbf{1})$ as in-degree matrix.
The $l$-th ($l\in\{1,\ldots,\lambda\}$) spectral retrieval layers can be formulated as 
\begin{equation}
\begin{split}
%    & V^l_u = \sigma (g_{\theta} \star_G H_u^l) =\sigma(
& V^l_u =\sigma(
 \theta^l_u( A_u + I)H_u^{l}), \\
% & V^l_d= \sigma(g_{\theta} \star_G H_d^l) =
& V^l_d =
 \sigma(\theta^l_d (A_d +I)H_d^{l}),
\end{split}
\end{equation}
where $V^l_u, V^l_d \in \mathbb{R}^{N\times F_T}$ is the upstream and downstream node embedding after $l$-th spectral retrieval layer. $H^l_u,H^l_d$ denotes the output from $(l-1)$-th SSRGC-block. 
In particular, $H^0_u=H^0_d=H_0$ is the temporal representations from the CBDDC-block. 
$\theta^l_u, \theta^l_d \in \mathbb{R}$ is learnable parameters of spectral retrieval layers.
% $\star_G$ with kernel $g_\theta$. 
$\sigma(\cdot)$ is the sigmoid activation. 
% Note that we do not directly calculate the summation of all the embedding of the layer like $V_u = \sum^L_{l=0} \sigma (g_{\theta} \star_G H_u^l)$ as embedding of current layer will be the input of next Spatial Retrieval layer.

\noindent\textbf{Spatial Retrieval Layer}.
Apart from nodes of the traffic network, road segments of traffic network also play essential roles in revealing traffic system status. 
Features of road segments such as length, geographical location, and the number of lanes, may greatly influence the adjacent nodes. 
% For example, the uncertainty of traffic flow increases as the road becomes longer due to the unexpected accidents on the road. 
However, recent studies~\cite{stgcn,astgcn,graphwavenet,stsgcn} mainly focus on extracting node embedding, instead of modeling the significance of the edges. 
Beyond these, locality and globality can also be simultaneously modeled when considering edges and its adjacent nodes. 
Globality represents the absolute flow of nodes since high flow indicates congestion, while low flow means the road segments are clear. 
% The link is seen as static from the perspective of globality. 
Locality reveals the transition volume of flows between the upstream and the downstream nodes in the near future. 
As the traffic network is highly dynamic, the spatial-correlation of the traffic network should be well captured through \emph{edge representations}.

Given an edge $s_1\to s_2$, we model the locality and globality by edge representation following spatial-based GCN. 
Specifically, the upstream edge representation $e_{u,s_1s_2}^l$ from node $s_1$ to node $s_2$ of $l$-th layer can be obtained by
\begin{equation}\label{eqn:uplink}
    e_{u,s_1s_2}^l=f_{\Theta^l_u}(V_{u,s_1}^l,V_{u,s_2}^l-V_{u,s_1}^l,W^1_{s_1s_2})
\end{equation}
where $f_{\Theta^l_u}(\cdot)$ is spatial retrieval layer with learnable parameter $\Theta^l_u$. 
The spatial retrieval layer is expected to fuse the features in global aspects as well as local aspects, where
$V_{u,s_1}^l$ is utilized to stand for the status of node $s_1$ from perspective of the global traffic network while $V_{u,s_2}^l-V_{u,s_1}^l$ represents the relative value of locality on edge $s_1\to s_2$. $W^1_{s_1s_2}$ denotes static edge features between $s_1$ and $s_2$.
With this design, globality, locality and the static edge information are incorporated via the spatial retrieval function $f_\Theta(\cdot)$. 

Similarly, the downstream edge representation $e_{d,s_2s_1}^l$ from $s_2$ to $s_1$ of $l$-th layer would be depected by: 
\begin{equation}\label{eqn:downlink}
    e_{d,s_2s_1}^l=f_{\Theta^l_d}(V_{d,s_2}^l,V_{d,s_1}^l-V_{d,s_2}^l,W^1_{s_2s_1}).
\end{equation}
% Note that the spatial retrieval layers share parameters to learn both upstream and downstream link embedding.

The edge representations are learned from both explicit and implicit features of the traffic network (RQ3).
As each sensor is usually connected to a small number of roads in real-world traffic network, the number of learned edge representation is in the scale of $\mathcal{O}(N)$.
To avoid further increase the complexity, we employ concatenation without parameters for both function $f_{\Theta^l_d}(\cdot)$ and $f_{\Theta^l_u}(\cdot)$.

\noindent\textbf{Aggregation Layer}.
Inspired by \cite{dgcnn} which chooses the k-nearest points of current node of the point cloud in 3D space and aggregates their information, we utilize an aggregation layer to amalgamate the edge representation. 
Intuitively, the most significant impact on node $s_1$ may come from its neighbor nodes with the shortest distance.
% For node $s_1$, the link that significantly impacts $s_1$ must be one adjoining $i$ and the distance between the other node $s_2$ on this link and $i$ should be smaller. 
In view of this rule, the aggregation layer to learn node embedding $H^{l+1}_{u,s_1}$ from upstream direction at $l+1$th block is designed as
\begin{equation}
    H^{l+1}_{u,s_1} = \bigcup_{s_2\in N_{d,k}(s_1)} \psi^l_{d,s_2} e_{u,s_1s_2}^l+\bigcup_{s_3\in N_{u,k}(s_1)}  \psi^l_{u,s_3} e_{u,s_3s_1}^l 
\end{equation}
where $\bigcup$ is an aggregation operation (e.g, summation or mean). 
$N_{u,k}(i)$ and $N_{d,k}(i)$ represent a set of $k$-nearest neighborhood node of node $i$ from upstream and donwstream seperately. 
$k$ can be set as the average degree of the traffic network. 
$\psi^l_{d,s}$ and $\psi^l_{u,s}$ are learnable parameters indicating the significance of nodes during the aggregation, and they are shared for all nodes in the traffic network for efficiency concern.
The node embedding $H^{l+1}_{d,s_1}$ from downstream direction can be defined in a similar way.

% In brief, the function of stacked SSRGC-block could be formulated as
% \begin{equation}
%   H = \|^{L-1}_{l=0} \bigvee(H^{l}_u,H^{l}_d;W) 
% \end{equation}
% where $\|$ denotes concatenation and $\bigvee(\cdot)$ represents the function of SSRGC-block as depicted in Eqn~(\ref{eqn:uplink}\--\ref{eqn:downlink}).
Finally, the outputs of each SSRGC-block are concatenated to capture both high-order features and low-order features.
Thus, globality and locality are both considered by TSSRGCN over the node embedding and edge representations (RQ2).
Specifically, we use a $1\times 1$ convolution layer to reduce the dimension to $F_S$. 
%The final output is denoted as $H$.

\subsection{Forecasting Layer}

We adopt skip connection \cite{resnet} to concatenate input $X$ with output of the aggregation layer $H$ as  input to the forecasting layer. 
To improve the efficiency, we directly use a fully connected layer to generate predicted value $\hat{Y}$ on each node at all $K$ time steps.
TSSRGCN is set to minimize the L2-loss between the predicted value $\hat{Y}$ and ground-true value $Y$, i.e., $\mathcal{L}(\hat{Y},Y) = \| \hat{Y} - Y\|^2$.

\section{Experiments}
% add experiments of Beijing highway, change pems to a larger dataset, add case study

In this section, we conduct experiments over real-world traffic datasets to examine the performance of TSSRGCN. We decomposite the RQs and design experiments to answer the following questions: \textbf{Q1}: How does our model perform compared to other state-of-the-art traffic flow forecasting models? \textbf{Q2}: Can our model capture the short-term and long-term temporal evolution patterns?

\subsection{Experiment Setup}
% \begin{table}[ht]
% \centering
% \caption{Hyperparameters of benchmark datasets}
% \label{tab:hypertab}
% \begin{tabular}{c|cc}
% \hline
%                 & PEMSD3 & PEMSD7 \\ \hline
% Learning Rate   & 0.01   & 0.001  \\
% $F_T$           & 64     & 64     \\
% $F_S$           & 64     & 64     \\
% $k$             & 3      & 3      \\
% $\lambda$       & 4      & 5      \\ \hline
% \end{tabular}
% \end{table}

\noindent\textbf{Datasets}.
 We evaluate our model on two real-world traffic datasets PEMSD3 and PEMSD7.
These datasets are collected by California Transportation Agencies (CalTrans) Performance Measurement System (PeMS) \cite{pems} by every 30 seconds, where the traffic flow around the sensors are reported (i.e., $F=1$).
The data are aggregated into every 5-minutes interval. 
% The system equips with more than 39,000 detectors placed on the highway in California's major metropolitan areas. 
The datasets also contain the metadata of the sensor network from which we can build the graph $G$. 

PeMSD3 contains data with 358 sensors in North Central Area from Sep. 1st to Nov. 30th in 2018. 
PeMSD7 contains data with 1047 sensors in San Francisco Bay Area from Jul. 1st to Sep. 30th in 2019.

\begin{table*}[tp]
\centering
\caption{Performance comparison of TSSRCNN and other baseline methods on PEMSD3.}
\label{tab:pemsd3}

\begin{tabular}{l|ccc|ccc|ccc}
\toprule
\hline
PEMSD3    & \multicolumn{3}{c|}{15 min} & \multicolumn{3}{c|}{30 min} & \multicolumn{3}{c}{60 min} \\ \hline
Model     & MAE     & RMSE    & MAPE($\%$)    & MAE     & RMSE    & MAPE($\%$)    & MAE     & RMSE    & MAPE($\%$)   \\ \hline
SVR & 18.28 & 29.98 & 24.44 & 21.00 & 33.66 & 26.10 & 24.33 & 38.87 & 29.46 \\
LSTM & 17.12 & 28.34 & 22.57 & 18.92 & 31.12 & 24.40 & 22.28 & 36.06 & 29.09 \\
DCRNN & 14.81 & 24.43 & 14.22 & 16.80 & 27.64 & 15.85 & 20.39 & 32.93 & 19.13 \\
STGCN & 14.78 & 27.15 & 21.45 & 16.83 & 29.79 & 24.32 & 20.59 & 34.93 & 28.19 \\
ASTGCN & 16.14 & 27.45 & 16.48 & 17.41 & 29.90 & 17.68 & 19.16 & 33.32 & 19.49 \\
Graph WaveNet & 14.61 & 24.89 & 15.00 & 16.50 & 28.11 & 15.68 & 20.12 & 33.38 & 18.32 \\
STSGCN & 14.82 & 23.92 & 14.74 & 15.81 & 25.64 & 15.52 & 17.61 & 28.69 & 16.95 \\\hline
\textbf{TSSRGCN} & \textbf{13.49} & \textbf{20.40} & \textbf{13.99} & \textbf{13.81} & \textbf{21.10} & \textbf{14.15} & \textbf{14.22} & \textbf{21.87} & \textbf{14.52} \\ \hline
\bottomrule
\end{tabular}
\end{table*}

\begin{table*}[tp]
\centering
\caption{Performance comparison of TSSRGCN and other baseline methods on PEMSD7}
\label{tab:pemsd7}
\begin{tabular}{l|ccc|ccc|ccc}
\toprule
\hline
PEMSD7    & \multicolumn{3}{c|}{15 min}    & \multicolumn{3}{c|}{30 min}     & \multicolumn{3}{c}{60 min}    \\ \hline
Model       & MAE     & RMSE   & MAPE($\%$)    & MAE     & RMSE    & MAPE($\%$)    & MAE     & RMSE    & MAPE($\%$)   \\ \hline
SVR & 21.94 & 35.13 & 12.64 & 25.33 & 39.90 & 13.94 & 31.10 & 48.42 & 16.92 \\
LSTM & 20.98 & 34.36 & 12.10 & 24.27 & 38.62 & 12.40 & 29.49 & 46.55 & 16.06 \\
DCRNN & 20.00 & 32.71 & \textbf{10.17} & 22.70 & 36.90 & 11.53 & 27.59 & 44.06 & 14.39 \\
STGCN & 20.25 & 32.12 & 10.23 & 23.39 & 36.42 & 11.60 & 29.32 & 44.61 & 14.28 \\
ASTGCN & 19.61 & 31.58 & 10.67 & 20.78 & 33.61 & 11.25 & 22.34 & 36.37 & 12.14 \\
Graph WaveNet & 20.36 & 33.18 & 10.67 & 23.28 & 37.75 & 12.36 & 28.56 & 45.54 & 15.09 \\
STSGCN & 20.03 & 31.79 & 10.54 & 21.33 & 33.88 & 11.14 & 23.57 & 37.43 & 12.25   \\\hline
\textbf{TSSRGCN} &\textbf{17.95} & \textbf{28.20} & 10.86 & \textbf{18.59} & \textbf{29.31} & \textbf{11.03} & \textbf{19.38} & \textbf{30.64} & \textbf{11.64} \\
\hline
\bottomrule
\end{tabular}
\end{table*}

\noindent\textbf{Preprocessing}.
The sampling frequency $T_{sp}$ is 5 minutes for two datasets, and there are 288 timesteps in one day.
The missing data are calculated by linear interpolation.
% Linear interpolation method is used to fill the missing data. 
Besides, the input data are transformed by zero-mean normalization. 
% as 
% \begin{equation}
%     \Tilde{X} = \frac{X-\mu}{\sigma}
% \end{equation}
% where $\mu$ and $\sigma$ are the mean and standard deviation of the raw data, respectively.

% The weighted adjacency matrices consist of adjacency matrix 
$W^0$ is adjacency matrix revealing real edges between nodes on the graph. $W^1$ is distance-base adjacency matrix which is defined as $W^1_{s_is_j}=\exp\left(-\frac{d_{s_is_j}^2}{\sigma_d^2}\right)$ if $ s_i \ne s_j$ and $ \exp\left(-\frac{d_{s_is_j}^2}{\sigma_d^2}\right) \ge \epsilon $. 
Here $d_{s_is_j}$ denotes the distance between sensors $s_i$ and $s_j$. $\sigma_d$ is the standard deviation of distance and  $\epsilon$ is 0.5 to control the sparsity of $W^1$ according to \cite{stgcn}.

\noindent\textbf{Settings}.
We implement TSSRGCN by PyTorch and select mean operation in the aggregation layer.
Adam \cite{adam} is leveraged to update the parameters during training for a stable and fast convergence.
The datasets are split into training/validation/test sets with ratio 6:2:2 in the time dimension.
Our task is to forecast traffic flow in the next hour as $K=12$.
We use the last hour data before the predicted time as the recent data, and the same hour of the last seven days to extract daily pattern and weekly pattern (daily pattern only requires the last three days' data). 
In this case, $\tau=\{1,12,84\}$ and we fix $P=3$ to capture the recent/daily/weekly patterns and the corresponding period shiftings. There are $T_N=96$ time steps in total.
% We use data of 8-day length to construct samples, which means $T_N=96$. 
% The period set $\tau=\{12, 7\times12\}$ extracts daily pattern and weekly pattern as long as a recent trend over the last 12 time steps. The task is to
 The batch size is 64 for PEMSD3 and 16 for PEMSD7 as the latter is about three times larger than the former. $F_T,F_S$ and $k$ is 64, 64 and 3 for both dataset. We set 1e-2 as the learning rate for PEMSD3 and 3e-3 for PEMSD7. $\lambda$ is 4 for PEMSD3 and 5 for PEMSD7.
 % Other hyperparameters can be found in Tab.~\ref{tab:hypertab}.

\noindent\textbf{Baseline Methods}.
We compare our model with the following baselines and we use the optimal hyperparameters of these methods mentioned in the corresponding paper:  \textbf{SVR} \cite{drucker1997support}, \textbf{LSTM} \cite{lstm},  \textbf{DCRNN} \cite{dcrnn}, \textbf{STGCN} \cite{stgcn}, \textbf{ASTGCN} \cite{astgcn}, \textbf{Graph WaveNet} \cite{graphwavenet}, 
    \textbf{STSGCN} \cite{stsgcn}.

\noindent\textbf{Evaluation Metric}.
To evaluate the performance of different models, we adopt Mean Absolute Errors (MAE), Root Mean Squared Errors (RMSE) and Mean Absolute Percentage Errors (MAPE) as our metric.

\subsection{Experimental Results (Q1 and Q2)}
We discuss the performace of TSSRGCN and other baselines, and compare the results over different time windows.

\noindent\textbf{Overall Performance (Q1)}.
We compare TSSRGCN with seven models on PeMSD3 and PeMSD7. 
Tab.~\ref{tab:pemsd3} and Tab.~\ref{tab:pemsd7} show the performance on the forecasting task at different time granularity (i.e., 15-, 30-, and 60-mins in the future). 
We can conclude that:
(1) Deep learning methods, especially models based on GCNs, perform better than traditional ones.
Due to the complex spatial-temporal correlation of traffic networks, traditional methods fail to capture the latent features of all nodes at all time steps. 
LSTM can extract some temporal information from the traffic data, which helps improve the prediction compared with SVR.
GCNs based models are compelling on mining graph structure data when solving our task, outperforming general deep learning model in many metrics. 
(2) TSSRGCN performs well on both datasets, verifying the robustness of our model to various traffic patterns and different scales of nodes in the graph. 

\noindent\textbf{Performance on Different Time Windows (Q2)}.
To show the ability to extract short and long term temporal information, we conduct TSSRGCN on various time windows. We can find that:
(1) TSSRGCN achieves state-of-the-art results in medium and long term (30 and 60 mins) as the long-period information is captured in the CBDDC block, indicating that periodic patterns contribute to extract temporal correlation. 
(2) TSSRGCN presents competitive performance to the best result on the prediction in the near future (i.e., 15-min), which can be attributed to the combination of both locality and globality.
(3) Models mining various temporal patterns can perform well both in the short and long term (i.e., ASTGCN and TSSRGCN).
%However, it performs not so successful on MAPE compared with our methods as our result has a significant improvement in it. We also achieve the second-best performance on MAE and RMSE, providing the evidence for the superiority of our model.

% \begin{figure}[t]
%     \centering
%     \includegraphics[width=\linewidth]{pic/Component_analysis_of_TSSRGCN_on_PEMSD3.png}
%     \caption{The ablation study of TSSRGCN on the PEMSD3 dataset. The word `lost' in the rear of the module indicates the exclusion of the corresponding module.}
%     \label{fig:without}
% \end{figure}
% \begin{figure}[h]
%     \centering
%     \subfigure[PEMSD3]{
%         \includegraphics[width=0.2\textwidth]{pic/pemsd3.PNG}
%     }
%     \subfigure[PEMSD7]{
%     \includegraphics[width=0.2\textwidth]{pic/pemsd7.PNG}
%     }
%     \caption{The distribution of sensors of datasets PEMhSD3 and PEMSD7.}
%     \label{fig.dataset}
% \end{figure}

	\section{Conclusion}
	In this paper, we propose TSSRGCN for traffic flow forecasting. 
	Motivated by the fact that there exist different temporal traffic patterns with period shifting, TSSRGCN employs the cycle-based dilated convolution blocks to incorporate the temporal traffic patterns from both short-term and long-term aspects.
	Meanwhile, GCNs for learning node embeddings and edge representations are stacked to retrieve spectral and spatial features from traffic network.
	Experiments on two real-world datasets show that our model performs well on different metrics compared to state-of-the-art methods, indicating robustness of the proposed method on various temporal patterns and the practicability to help administrator regulate the traffic in the real world. 
\bibliographystyle{IEEEtran}



% \end{thebibliography}
\bibliography{ref}

\end{document}